\documentclass{article}

\usepackage[preprint,nonatbib]{neurips_2021}
\usepackage{graphicx}
\usepackage{caption}

\usepackage[utf8]{inputenc} 
\usepackage[T1]{fontenc}    
\usepackage{hyperref}       
\usepackage{url}            
\usepackage{booktabs}       
\usepackage{amsfonts}       
\usepackage{nicefrac}       
\usepackage{microtype}      
\usepackage{xcolor}         

\usepackage{amsmath}
\usepackage{booktabs} 
\usepackage{array}

\DeclareMathOperator*{\argmax}{argmax}

\newcolumntype{F}[1]{
    >{\raggedright\arraybackslash\hspace{0pt}}m{#1}}
\newcolumntype{T}[1]{
    >{\centering\arraybackslash\hspace{0pt}}m{#1}}

\title{Encoders and Ensembles for Task-Free Continual Learning}

\author{
  Murray Shanahan \\
  DeepMind and Imperial College London\\
  \texttt{mshanahan@deepmind.com} \\

  \And
  Christos Kaplanis \\
  DeepMind \\
  \texttt{kaplanis@deepmind.com} \\
  \And
  Jovana Mitrovi{\'c} \\
  DeepMind \\
  \texttt{mitrovic@deepmind.com} \\
}

\begin{document}

\maketitle

\begin{abstract}
We present an architecture that is effective for continual learning in an especially demanding setting, where task boundaries do not exist or are unknown, and where classes have to be learned online (with each example presented only once). To obtain good performance under these constraints, while mitigating catastrophic forgetting, we exploit recent advances in contrastive, self-supervised learning, allowing us to use a pre-trained, general purpose image encoder whose weights can be frozen, which precludes forgetting. The pre-trained encoder also greatly simplifies the downstream task of classification, which we solve with an ensemble of very simple classifiers. Collectively, the ensemble exhibits much better performance than any individual classifier, an effect which is amplified through specialisation and competitive selection. We assess the performance of the encoders-and-ensembles architecture on standard continual learning benchmarks, where it outperforms prior state-of-the-art by a large margin on the hardest problems, as well as in less familiar settings where the data distribution changes gradually or the classes are presented one at a time.
\end{abstract}

\section{Introduction}

Supervised learning methods often rely on the assumption that data is identically and independently distributed (i.i.d.) and drawn from a fixed distribution. However, in many applications, as in ordinary human life, this assumption is unwarranted. Instead, data arrives a little at a time from an ever-changing world upon which the learning system has a limited window. The sub-field of {\em continual learning} studies algorithms and architectures that can cope with this more realistic scenario \cite{hadsell2020embracing, delange2021continual}. The main difficulty to be overcome in devising neural network architectures for continual learning is that of {\em catastrophic forgetting}, wherein a model's performance on earlier tasks in a sequence degrades as it learns new tasks \cite{mccloskey1989catastrophic, french1999catastrophic, kirkpatrick2017overcoming}. Neural networks are especially vulnerable to catastrophic forgetting because of the holistic nature of gradient-based weight updates \cite{hadsell2020embracing}. Each new example that a network encounters has the potential to perturb every weight in the network. If a whole class of examples is absent from the data for a while, weights tuned for that class soon drift. Researchers have addressed this problem in a number of ways, such as consolidating weights, retaining a memory of past experience, dividing the architecture into separate modules, and meta-learning \cite{hadsell2020embracing, delange2021continual}.

Here we demonstrate that catastrophic forgetting can be mitigated, and current state-of-the-art results surpassed, in the most challenging continual learning setting, using a relatively straightforward combination of well-understood architectural components (Fig.~\ref{fig:architecture}). This is possible largely thanks to recent progress in self-supervised learning. Our architecture incorporates a powerful off-the-shelf encoder whose self-supervised pre-training has resulted in a general-purpose feature extractor \cite{grill2020bootstrap, mitrovic2021representation}, which greatly simplifies the downstream classification task. (The encoder must be pre-trained on a dataset different from the one used to evaluate continual learning performance, of course.) By freezing the weights of the pre-trained encoder, we ensure that this part of the architecture is immune to forgetting. However, forgetting can still creep in downstream of the encoder. We address this by using an ensemble of very simple single-layer classifiers. Individually, these classifiers are prone to errors, but collectively they perform well. Their collective success is amplified through specialisation, which is achieved by associating each classifier with a random key drawn from the encoder’s latent space, and by selecting a subset of classifiers according to key-matching. Finally, a judicious choice of activation and loss functions further mitigates forgetting by confining updates to weights that affect the class of the current sample.

Much of the continual learning literature makes the unrealistic assumption that the learning problem is constituted by a set of sequentially presented tasks, each comprising data drawn from a different distribution. Indeed, many approaches rely on the existence of well defined task boundaries within the data. By contrast, the most realistic continual learning setting, as well as the most challenging, is one in which no clear task boundaries even exist \cite{aljundi2019task, aljundi2019online}. This is the task-free continual learning setting we tackle in the present paper (although, for convenience, we still refer to tasks where appropriate). The architecture we present here does not require knowledge of task boundaries, nor does it try to infer them, and it can be applied to continual learning problems where the distribution changes gradually and no clear task boundaries exist. We evaluate the architecture on three standard image classification datasets, namely MNIST, CIFAR-10, and CIFAR-100. We look at three benchmark variations, each tailored for continual learning: 1) task-wise splits, where the classes are partitioned into subsets (tasks) and presented sequentially, 2) the fully incremental case, where the classes are presented one at a time, and 3) a variation in which the classes are drawn from a distribution that shifts gradually over the course of training, presenting an ever-changing blend of classes.

\section{An Ensemble Memory Architecture}

Our architecture comprises a pre-trained image encoder with an ensemble of single-layer classifiers (Fig.~\ref{fig:architecture}). (For full details, including hyperparameters, see the Supplementary Material.) Each classifier has a fixed associated key, which is used for classifier look-up via $k$-nearest neighbours. Hence the ensemble can be thought of as residing in a memory. Crucially, the key space of the ensemble memory is the same as the latent space of the encoder. The architecture works as follows. An image is passed through the encoder, and the resulting encoding is used to look up the $k$ classifiers with the closest keys by cosine similarity. The encoding is then passed through each of these classifiers, and the resulting vectors are aggregated to produce the model's final output. The classifiers' output vectors are aggregated by taking their weighted average, where the weighting for each classifier is given by the cosine similarity between the encoding and the classifier's key. As we shall see, to mitigate catastrophic forgetting, the loss function and the classifiers' activation functions must be of a specific form.

\begin{figure*}[t]
  \centering
  \includegraphics[width=0.9\textwidth]{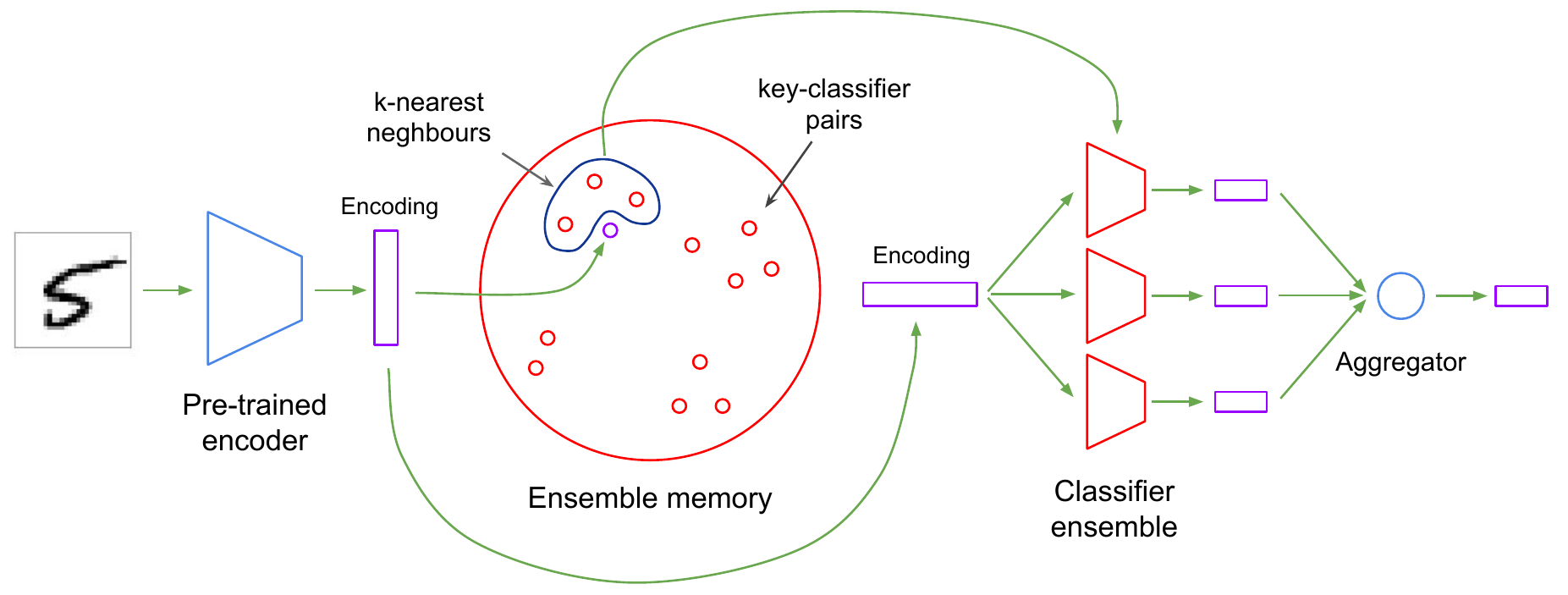}
  \caption{The Ensemble Memory Architecture}
  \label{fig:architecture}
\end{figure*}

Let $x$ be an input image and $y \in \{0,1\}^m$ be the one-hot encoding of the class to which it belongs. Let $f$ be an encoder with $z=f(x)\in \mathbb{R}^{d}$ the latent encoding. 
Let a {\em t-classifier} ($\mathrm{tanh}$ classifier) be a pair $(W,b)$ where $W \in \mathbb{R}^{m \times d}$ is a matrix of trainable weights and $b \in \mathbb{R}^m$ is a vector of trainable biases. The output of a t-classifier is given by
\begin{align}
\label{equation:t_classifier}
v(W,b,z) & = [\phi(\psi_1(z)), \dots, \phi(\psi_m(z))]
\end{align}
where $\psi_i(z) = w_i \cdot z^{T} + b_i$ and $\phi(x) = \tau \tanh(x / \tau)$ with $w_{i}$ the $i^{\text{th}}$ row of $W$. The scaling factor $\tau$ is a hyper-parameter. Now, let an ensemble memory of size $n$ be a pair $M = (M_{\text{key}}, M_{\text{cfier}})$ of vectors of keys and t-classifiers, respectively, i.e. $M_{\text{key}}\in\mathbb{R}^{n\times d}$ and $M_{\text{cfier}}\in(\mathbb{R}^{n\times m\times d}, \mathbb{R}^{n\times m})$ representing the weights and biases of the t-classifiers. 
We denote the $i^{\text{th}}$ key in $M_{\text{key}}$ and the $i^{\text{th}}$ classifier in $M_{\text{cfier}}$ by $M_{\text{key}}^i$ and $M_{\text{cfier}}^i$, respectively. Let $\gamma(x,y)$ be the cosine similarity between two vectors $x$ and $y$, and let $\mathcal{I}(i,z)$ denote the index of the $i^{\text{th}}$-ranked key in $M_{\text{key}}$ according to cosine similarity to the encoding $z$. Then the output of the model is given by
\begin{align}
V_M(z) = \frac{\sum_{i=1}^k \gamma(M_{\text{key}}^{\mathcal{I}(i,z)},z) v(W^{\mathcal{I}(i,z)}, b^{\mathcal{I}(i,z)},z)}{\sum_{i=1}^k \gamma(M_{\text{key}}^{\mathcal{I}(i,z)},z)} 
\end{align}
where $M_\text{cfier}^{\mathcal{I}(i,z)}=(W^{\mathcal{I}(i,z)}, b^{\mathcal{I}(i,z)})$. We seek to minimise the loss function
\begin{align}
\mathcal{L}(y, \hat{y}) & = -(y \cdot \hat{y})
\end{align}
where $\hat{y}=V_M(f(x))$ (the predicted labels from the model).

As hinted at earlier, several unconventional features of the model are essential to its success. First, the loss function is simply the dot product of the model's output with a one-hot vector, with no softmax or other form of normalisation applied. Second, the activation function for a t-classifier is $\tanh$ with a scaling factor $\tau$ applied, which allows the output of a neuron to grow ever closer to $\tau$ without ever reaching it. A third unconventional feature of our approach is the choice of optimiser. We found training to be most effective when, for each batch, the magnitude of a parameter's gradient was discarded having determined its sign, so that each parameter is raised or lowered by a fixed step size (the learning rate).

The pre-trained encoder $f$ is an essential component of the architecture's design. Here we employ two kinds of encoder. For the easier MNIST dataset, a simple variational autoencoder (VAE) is sufficient \cite{kingma2014autoencoding, rezende2014stochastic}. The encoder half of our VAE comprises two convolutional layers followed by two linear layers, while the mirror-image decoder comprises two linear layers followed by two transpose convolutional layers. Crucially, for the architecture to count as performing continual learning, the encoder must be pre-trained on a {\em different dataset} to the one used during the continual learning phase itself. In our case, for evaluation on MNIST, we pre-trained the VAE to reconstruct Omniglot characters \cite{lake2015human}, then discarded the decoder half. The encoder half, with its weights frozen, was retained for the full model.

For more challenging datasets than MNIST, a more capable pre-trained encoder is needed. Recently there has been considerable progress in the use of self-supervised contrastive learning to pre-train image encoders \cite{grill2020bootstrap, chen2020simple, mitrovic2021representation}. We obtained results using both ReLIC \cite{mitrovic2021representation} and BYOL \cite{grill2020bootstrap}, in each case with a ResNet-50 encoder \cite{he2016deep} pre-trained on the ImageNet dataset \cite{deng2009imagenet}. Best performances were obtained using ReLIC, where we closely followed the setup in \cite{mitrovic2021representation}, using the same hyperparameters and optimisation settings. The results obtained with BYOL were also good, although not quite as impressive as with ReLIC (see Table~\ref{table:byol_results}). By leveraging a comparison to both similar and dissimilar points, ReLIC learns a latent space in which representations are more tightly clustered according to the latent classes, whereas BYOL only focuses on a comparison to similar points and thus learns a latent space which is less tightly clustered.

\section{Evaluation}

\textbf{Benchmarks}. We evaluated our architecture on three standard image classification datasets: MNIST, CIFAR-10, and CIFAR-100. For MNIST and CIFAR-10, we trained models on three continual learning benchmarks: 5-way split, 10-way split (one class at a time), and a Gaussian schedule (Fig.~\ref{fig:gaussian_schedule}). (For full details of each protocol, see the Supplementary Material.) In the conventional (i.i.d.) setting, a model is presented with samples drawn uniformly from every class in the training set throughout training, but in these continual learning benchmarks, samples are drawn from a subset of classes that changes as training proceeds. In the 5-way split, the class labels are partitioned into five subsets of two labels each (five tasks) and the model is presented with one subset at a time. Confusingly, the term ``split MNIST’’ has been used by different authors to designate different protocols. As Aljundi, {\it et al.} point out, ``comparing reported results in continual learning requires great diligence because of the plethora of experimental settings’’ \cite{aljundi2019online}. For example, a number of authors (eg: \cite{zenke2017continual, farajtabar2020orthogonal}) use the term ``split MNIST'' to designate a protocol in which models are tested on five binary classification tasks, where the first task is to distinguish between digits 0 and 1, the second task is to distinguish between digits 2 and 3, and so on. By contrast, following Aljundi, {\it et al.} \cite{aljundi2019online} and Lee, {\it et al.} \cite{lee2020neural}, for our 5-way split benchmark, we calculate accuracies for 10-way classification, since there are 10 classes in the dataset. Hsu, {\it et al.} \cite{hsu2018reevaluating} provide a clear account of these different evaluation protocols. (See also \cite{vandeven2019, prabhu2020gdumb}.) According to their taxonomy, we perform ``incremental class learning’’ in the 5-way split benchmark, also known as the ``single head'' or ``shared classifier'' setting \cite{vandeven2019}, which, as they show, is the most difficult.

The other two benchmarks we use are of a less common type. In the {\em fully incremental} 10-way split, the model receives samples from one class label at a time. To be successful at this benchmark, the model has to learn the first class having never seen examples from any other class, and then to correctly re-identify members of that class throughout training. The 5-way and 10-way split benchmarks both feature distinct tasks and clear task boundaries. However, our model makes no use of this fact, and works in the more natural task-free setting. The {\em Gaussian schedule} benchmarks test for this capability. A {\em schedule}, in this context, defines how a data distribution evolves over training. In the Gaussian schedule, there are no task boundaries. Rather, each label appears in the data with a probability that follows a bell curve, and each label's probability peaks at a different time (see Fig.~\ref{fig:gaussian_schedule}). In our case, the probability peaks corresponding to the ten classes are evenly spread out over training, and the respective curves overlap significantly, so that one class blends smoothly into the next. For the CIFAR-100 dataset, we trained the model on three benchmarks: 20-way split, 100-way split, and a Gaussian schedule. In the 20-way split benchmark, the dataset's 100 classes are partitioned into 20 subsets of five classes each, according to their coarse class labels. In the fully incremental 100-way split, the classes are presented and learned one at a time, analogously to 10-way split MNIST and 10-way split CIFAR-10. For both benchmarks, accuracies are calculated for 100-way classification.

\begin{table}

  \caption{Accuracy (\%) (higher is better). Results for GEN-MIR and ER-MIR are taken from \cite{aljundi2019online}. Results for CN-DPM are taken from \cite{lee2020neural}. Results for GDumb are taken from \cite{prabhu2020gdumb}. For our models, means and standard deviations for 20 runs are shown.}
  \label{table:accuracy}
  
  \vspace{0.2cm}
  
  \centering
    \includegraphics[width=1.0\textwidth]{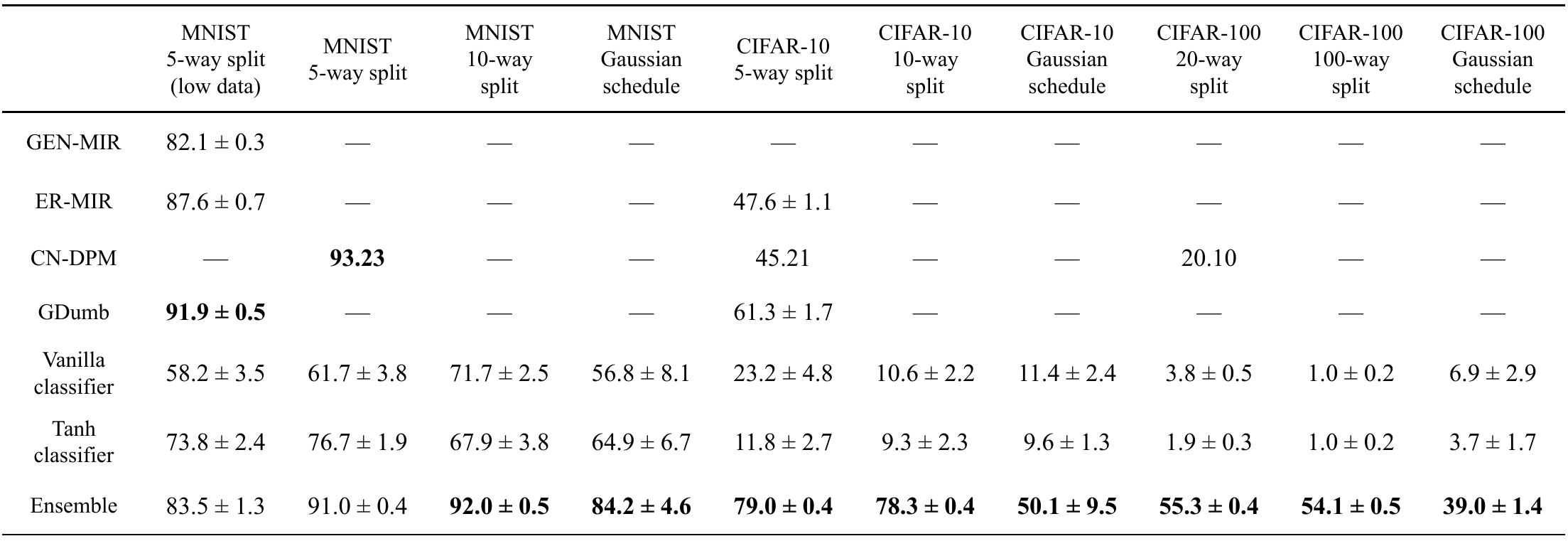}
  
\end{table}

\textbf{Findings}. Our main findings are summarised in Table~\ref{table:accuracy}, alongside previous state-of-the-art performances, where applicable. To reiterate, all final accuracies are for $n$-way classification, where $n$ is the total number of class labels. So $n=10$ for MNIST and CIFAR-10, while $n=100$ for CIFAR-100. The models we compare with are Maximal Interfered Retrieval (MIR) technique of Aljundi, {\it et al.} \cite{aljundi2019online}, the Continual Neural Dirichlet Process Mixture (CN-DPM) method of Lee, {\it et al.} \cite{lee2020neural}, and the Greedy Sampler and Dumb Learner (GDumb) method of Prabhu {\it et al.} \cite{prabhu2020gdumb}. MIR is a replay-based method that impoved on prior art, such as \cite{lopezpar2017gradient}, by selectively, rather than randomly, sampling from memory. It comes in two variants: GEN-MIR, which samples from encodings produced by a generative model, and ER-MIR, which directly samples stored images. CN-DPM is a mixture-of-experts style approach, where the set of ``experts'' (Dirichlet processes) expands dynamically to accommodate shifts in data distribution. GDumb is another a replay-based method, but unlike methods that mitigate forgetting by fine-tuning on stored samples, in GDumb the whole model is re-trained from scratch on all the stored samples at inference time. The precludes forgetting by design, but at the cost of inference-time computation.

Following Aljundi, {\it et al.} and Lee, {\it et al.}, our training regime is ``online’’ in the sense that it does not involve multiple passes through the data (multiple epochs). This contrasts with, for example, iCaRL, whose reported results were obtained with multiple epochs \cite{rebuffi2017icarl, vandeven2019}. Where Aljundi, {\it et al.}, Lee, {\it et al.}, and Prabhu {\it et al.} use a common protocol (5-way split CIFAR-10), we followed suit to obtain our results. Where they use different protocols (5-way split MNIST), we produced results for the protocols used by each set of authors (ie: for both a low data regime and for the full training set). In the case of CIFAR-100, only Lee, {\it et al.} supply a previous result. It should be born in mind that, as far as methods are concerned, we are not comparing like-with-like. For example, our method exploits the availability of a pre-trained encoder, unlike the other methods, while GDumb has a large computational overhead at inference time, unlike other methods (including ours). Claims of state-of-the art should be taken in this context. The ultimate value of the comparison is to contribute to a map of design possibilities for architectures for continual learning.

For 5-way split MNIST, our ensemble model's performance is slightly below the previous best accuracy with the full training set (CN-DPM), and falls between GEN-MIR and ER-MIR for the low data regime. The best performance in for MNIST in this setting is obtained by GDumb, albeit with a heavy price tag in terms of computation at inference time. We note that MNIST is an ``easy'' dataset for which state-of-the-art accuracies are already high, and the pre-trained encoder we used to obtain MNIST results was a simple VAE. Our real focus here is the sort of realistic colour images exemplified by the CIFAR datasets, for which current state-of-the-art accuracies remain low. This is where powerful encoders pre-trained using contemporary self-supervised methods come into their own. For 5-way split CIFAR-10, the ensemble model achieves a final accuracy of 79.0\%, which improves on ER-MIR (state-of-the-art in 2019) by over 30\% and on GDumb by 18\% (previous best reported accuracy). Finally, for 20-way split CIFAR-100, the most demanding dataset, our model achieves a final accuracy of 55.3\%, which also represents an improvement on prior state-of-the-art (CN-DPM) by over 30\%. We additionally report accuracies for the fully incremental (10-way or 100-way) split for each of MNIST, CIFAR-10, and CIFAR-100, as well as accuracies for the Gaussian schedule for each dataset. The performance of the ensemble model on the 10-way split is within 1\% of its performance on the 5-way split for both MNIST and CIFAR-10. For 100-way CIFAR-100, the ensemble model attains 54.1\%, which is close to its performance on the 20-way split. The Gaussian schedule presents more of a challenge for each dataset, although in every case the ensemble method is the best performer.

Overall, our findings show that the encoders-and-ensembles approach is effective for task-free continual learning, and that it outperforms other models on harder datasets. We note that, in addition to the differences between methods mentioned above, we have not compared like-with-like in terms of memory requirements; our model incorporates an encoder with a large number of parameters, and includes a memory-hungry ensemble. That said, even a small ensemble is sufficient to achieve state-of-the-art results (see ablations below), and replay-based methods also have extra memory requirements, thanks to their need for a replay buffer. Moreover, in other branches of machine learning, progress is often made by building bigger models, reflecting the ever-increasing memory capacity of commodity hardware. Although optimising memory usage is clearly a worthwhile aim, the continual learning sub-field also needs to explore ways to make good use of more memory when it is available.

\textbf{Forgetting}. In addition to classification accuracy, we assessed each model's propensity to forget what it has learned during the course of training. To do this, we generalise the measure defined in \cite{chaudhry2018riemannian} to cover the task-free setting. We define the {\em generalised forgetting} of a model over a training period with $n$ time steps (batches) as
\begin{align}
    \frac{1}{m}\sum_{i=1}^{m} \max\limits_t (a^{i}_{t} - a^{i}_{n})
\end{align}
where $m$ is the number of classes and $a^{i}_{t}$ is the accuracy of the model for class $i$ at time $t$. The forgetting metrics for our experiments are reported in Table~\ref{table:forgetting}. As with accuracy, we show results for previous state-of-the-art \cite{aljundi2019online} where applicable. (We note that generalised forgetting, which we report for our model, is a strict upper bound on task-wise forgetting, as reported in \cite{aljundi2019gradient}. Lower forgetting is better, of course.) We found that the ensemble model outperforms the state-of-the-art where previous figures were reported. On 5-way split MNIST, our model yields 6.2\% (generalised) forgetting (albeit with high variance), compared to 7.0\% forgetting for ER-MIR, while on 5-way split CIFAR-10, our model yields 7.5\% compared to 17.4\% for ER-MIR.

\begin{table}

  \caption{Forgetting (lower is better). Results for GEN-MIR and ER-MIR are taken from \cite{aljundi2019online}. Means and standard deviations for 20 runs are shown.}
  \label{table:forgetting}
  
  \vspace{0.2cm}
    
  \centering
    \includegraphics[width=1.0\textwidth]{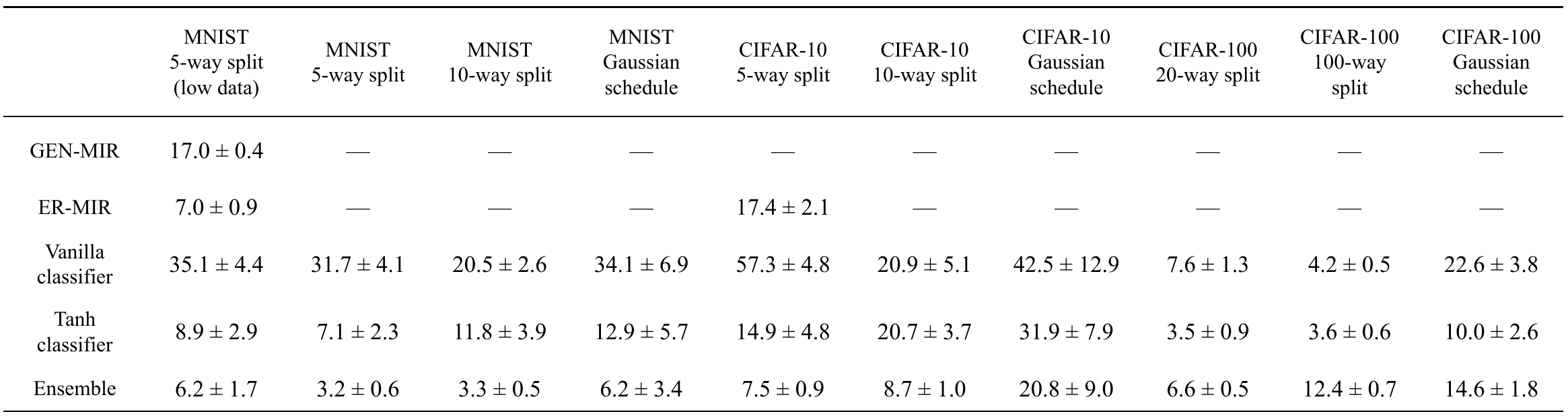}

\end{table}

\subsection{Baselines and Ablations}

\begin{figure*}
  \centering
  \includegraphics[width=1.0\textwidth]{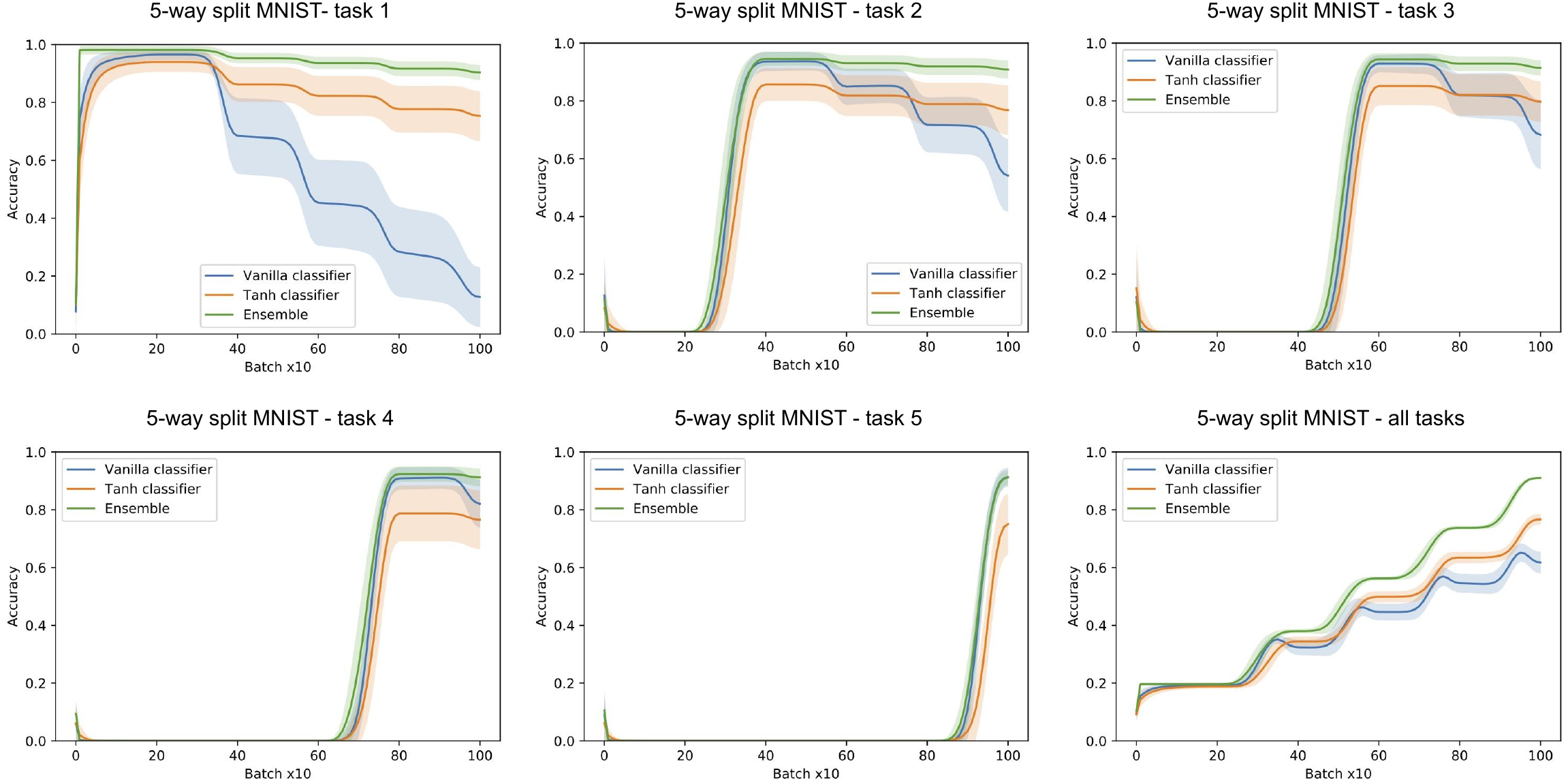}
  \caption{Task-wise accuracies over training for 5-way split MNIST}
  \label{fig:mnist-results}
\end{figure*}

\begin{figure*}
  \centering
  \includegraphics[width=1.0\textwidth]{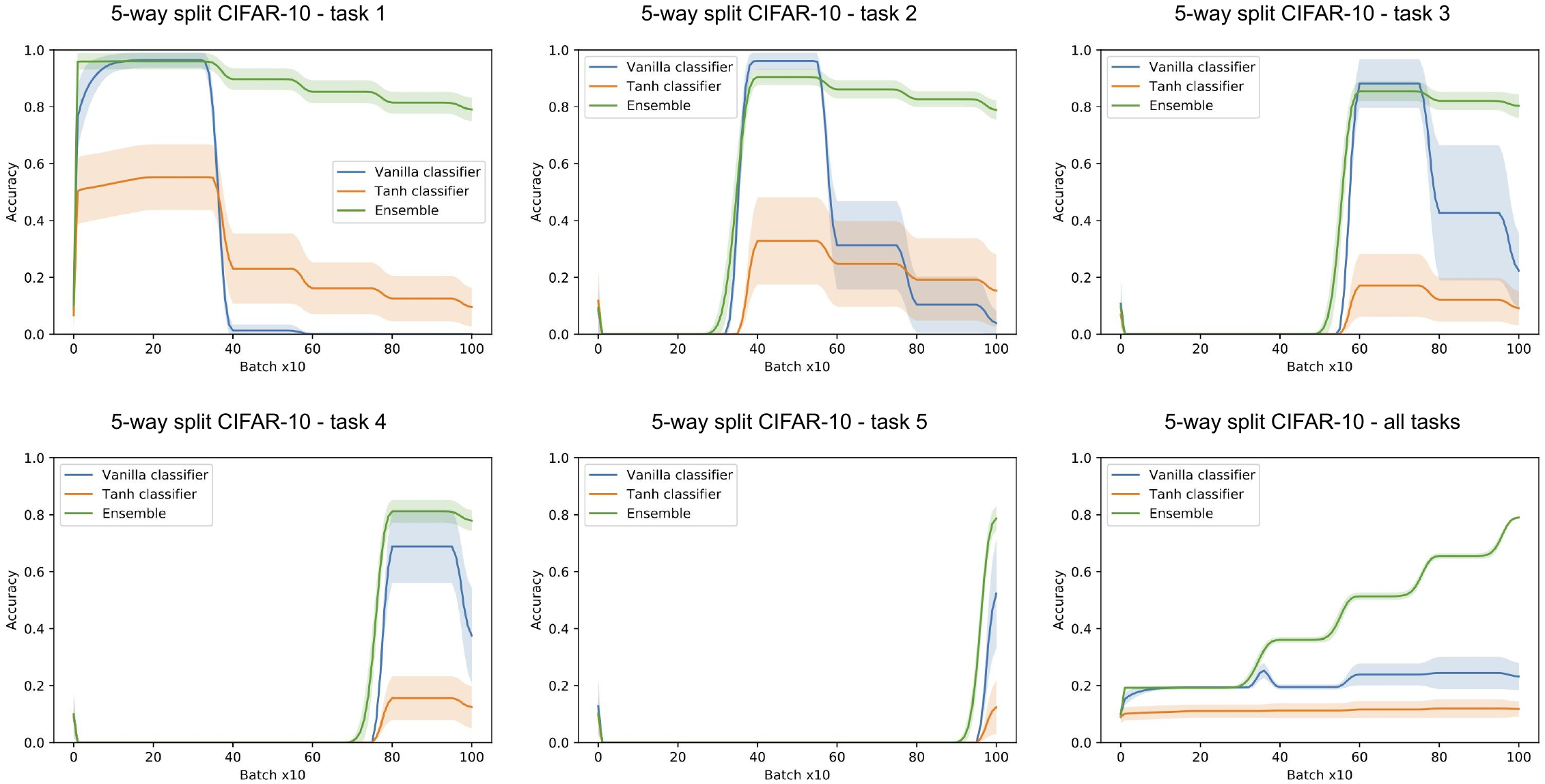}
  \caption{Task-wise accuracies over training for 5-way split CIFAR-10}
  \label{fig:cifar10-results}
\end{figure*}

For a more qualitative appreciation of the ensemble model in action, see Fig.~\ref{fig:mnist-results}, bottom right, and Fig.~\ref{fig:cifar10-results}, bottom right, which show the evolution of overall accuracy on all ten classes on the test set for 5-way split MNIST and 5-way split CIFAR-10, respectively. The model's progress is steadily upwards, and where there are discrete changes in distribution (new tasks) -- in the 5-way and 10-way splits -- their effects are clearly visible. These figures also show the performance of two non-ensemble baselines models, one in which the output of the pre-trained encoder is delivered to a single vanilla classifier with a conventional softmax applied, and a similar model with a single t-classifier (with a $\mathrm{tanh}$ activation function and no softmax). (See Table~\ref{table:accuracy} for numerical final accuracies.) For the easier MNIST dataset (Fig.~\ref{fig:mnist-results}), forgetting is non-catastrophic even with the vanilla classifier, and is further reduced with the stand-alone $\mathrm{tanh}$ classifier, although neither model matches the performance of the ensemble. This suggests that each of the three elements of the architecture -- the pre-trained encoder, the activation / loss function combination, and the ensemble -- can help to reduce catastrophic forgetting for a simple enough dataset. However, with the harder CIFAR-10 dataset (Fig.\ref{fig:cifar10-results}), catastrophic forgetting is mitigated only with the full ensemble model, incorporating all three of these features.

The remaining plots in Figs.~\ref{fig:mnist-results} and \ref{fig:cifar10-results} depict task-wise accuracies for 5-way split MNIST and 5-way split CIFAR-10, respectively. The propensity for forgetting in the vanilla classifier is clear, although it is more rapid for CIFAR-10 than MNIST. This contrasts with the ensemble model, which exhibits much less forgetting with both datasets. The stand-alone $\mathrm{tanh}$ classifier exhibits markedly less forgetting than the vanilla classifier for MNIST, but it attains a worse initial performance on each task, putting its final accuracy between that of the vanilla classifier and the ensemble model. A similar trade-off is apparent for CIFAR-10 with the stand-alone classifier, where its initial performance on each task is especially poor.

We carried out a number of further experiments to assess the extent to which ensemble size and top-$k$ classifier selection contribute to the model's success. We found that the ensemble model is still effective even with a significantly reduced ensemble size, though its performance degrades (Table~\ref{table:ablations}, left). Even with a ``small'' ensemble (128 classifiers, $k=8$), the model still beats ER-MIR on 5-way split CIFAR-10 in terms of final accuracy by 17\%, and is comparable to GDumb, which has large overheads in terms of both memory and inference-time computation. This is encouraging, as an increase in memory requirements is one of the costs of our approach. The best performance, however, is still acheieved with the largest ensemble (1024 classifiers), as shown in Table~\ref{table:accuracy}. To study the impact of top-$k$ classifier selection, we looked at the effect of larger values for $k$. We found that performance slowly degrades for $k>32$ (Table~\ref{table:ablations}, right), which suggests that the competition and consequent specialisation induced by top-$k$ selection is working as expected.

\section{How Does the Model Work?}

\begin{figure*}
  \centering
  \includegraphics[width=1.0\textwidth]{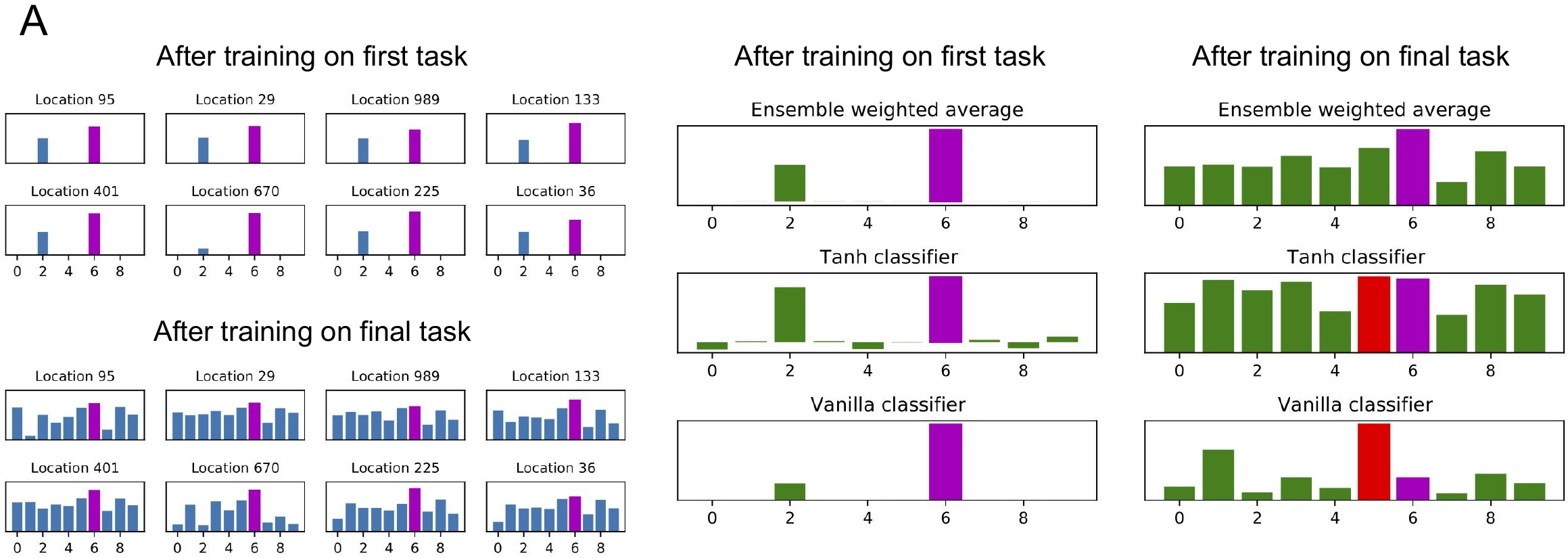}
  
  \vspace{0.2cm}
  
  \includegraphics[width=1.0\textwidth]{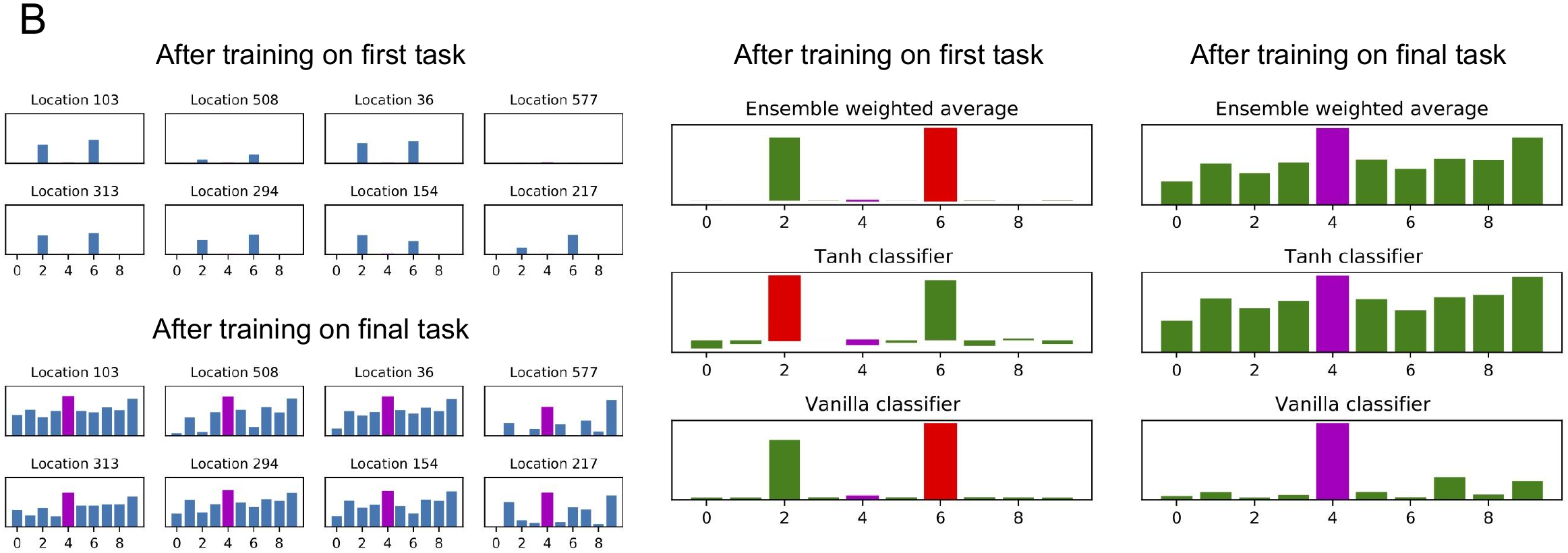}
  \caption{Visualisation of a representative training run for 5-way split MNIST (low data regime). Final layer activations are shown for two sample digits at different points in training. Magenta denotes the correct digit. Red denotes the maximal activation where different from the correct digit. (A): The sample digit (8) belongs to the first task. (B): The sample digit (3) belongs to the final task.}
  \label{fig:visualisation}
\end{figure*}

Here we offer some intuition for how the model works (visualised in Fig.~\ref{fig:visualisation}). Given a pre-trained encoder with fixed weights, the success of the model depends on two further features that operate in tandem: the dot-product loss function / $\text{tanh}$ activation function combination, and the ensemble. Let's first consider a single t-classifier. How, in the context of a fixed encoder, does the combination of a dot-product loss function and a $\text{tanh}$ activation function mitigate catastrophic forgetting? Let $f$ be the frozen, pre-trained encoder, and let $v_i$ be the $i^\mathrm{th}$ neuron in the t-classifier $v$ (representing class $i$). (So $v_i(x) = \phi_i(\psi_i(x))$ (Eqn.~\ref{equation:t_classifier}).) Let's consider $E(i,j) = \mathbb{E}(v_i(f(x)) | x \in C_j)$, the expected value output by neuron $i$ given an input image $x$ from class $C_j$. How does this expectation evolve during training? Because of the form of the loss and activation functions, given an image from class $j \neq i$, the gradients of the weights on incoming connections to $v_i$ are always zero. So incoming images from any class $j \neq i$ will not affect those weights, and therefore will not affect $E(i,j)$, and we can safely ignore this case. But what about incoming images from class $i$ itself? Given an image from class $i$, the weights on incoming connections to $v_i$ will be adjusted to push $v_i(f(x))$ up, which will drive up $E(i,i)$ over time. Let's call these {\em targeted} increases in $E$. Targeted increases in $E(i,i)$ are good, as they steer the model towards correct classifications. However, as a side-effect of increasing $E(i,i)$, $E(i,j)$ where $j \neq i$ can also increase. Let's call these {\em collateral} increases in $E$. Collateral increases in $E(i,j)$ where $j \neq i$ are bad, as they draw the model away from correct classifications.

Crucially, over time, as long as the data is roughly class-balanced, any collateral increases in $E(i,j)$ where $j \neq i$ will tend to be outpaced by the targeted increases in $E(i,i)$. Consequently, $E(i,i) - E(j,i)$ will tend to increase for all $i$ and $j$, which is what we really care about since this is what determines the predicted class after taking $\argmax_i v_i(f(x))$. Now, the tendency for targeted increases in $E$ to outpace collateral increases in $E$ is sometimes enough for a single t-classifier to mitigate catastrophic forgetting, as we see with the MNIST benchmarks in Fig.~\ref{fig:mnist-results}. However, as this desirable upward trend is rather weak, it isn't sufficient to prevent poorly performing t-classifiers from arising, as they do most of the time for the more challenging CIFAR-10 and CIFAR-100 benchmarks. This is where the ensemble comes in. The weak statistical trend we're looking for can be amplified by training many t-classifiers. Averaged over a population of t-classifiers, the tendency for targeted increases to outpace collateral increases will be greater than for a single t-classifier.

Moreover, an ensemble approach allows for specialisation, which further improves performance. In our architecture, this is achieved through top-$k$, key-based classifier lookup, where they keys reside in the same space as the image encodings. Any degree of class-relevant clustering in the encoder's latent space will confer an advantage here, further magnifying the statistical trend described above. To see this, recall that we are interested in increases in $E(i,i) - E(j,i)$ for all $i$ and $j$. We have already seen that,  for a standalone t-classifier, the targeted increases in $E(i,i)$ will tend to increase faster than the collateral increases in $E(i,j)$. Now consider a t-classifier from the ensemble. Assuming the latent space exhibits a degree of class-relevant clustering, the chances of the same classifier appearing in the top-$k$ set for two given images will be higher for images belonging to the same class than for images from different classes. This entails that the classifier will tend to be updated more often through images of some classes than others. Now suppose the classifier is more likely to be updated by images from class $i$ than class $j$. This entails that not only will the targeted increases in $E(i,i)$ tend to be larger per training step than the collateral increases in $E(i,j)$, they will also be more frequent. We further enhance this effect by taking the weighted average of the output of the top-$k$ classifiers, giving more weight to classifiers whose keys are close to an image’s encoding, in other words to those that are more likely to have received more training on similar images.

\section{Related Work}

\textbf{Ensembles}. The literature on continual learning is extensive (see \cite{hadsell2020embracing} and \cite{delange2021continual} for recent reviews), so we focus here on related work that either a) uses techniques similar to ours (especially ensemble methods), or b) addresses the task-free scenario. First, we consider ensemble methods, which have been studied widely within machine learning \cite{sagi2018ensemble}. Ensembles of one sort or another feature in several prior approaches to continual learning. For example, a number of authors have applied mixture of experts models \cite{jacobs1991adaptive} to continual learning, which explicitly incorporate ensembles \cite{aljundi2017expert, kruszewski2021evaluating, lee2020neural}. Others use ensembles in a more implicit way, in the form of sub-networks, to improve continual learning. For example, in \cite{fernando2017pathnet}, a genetic algorithm is used to discover an effective sub-network for a task, which is then frozen and can be reused when training on the next task. In other work, dropout \cite{srivastava2014dropout}, which creates implicit sub-networks by silencing neurons at random during training, has been shown to mitigate catastrophic forgetting in a continual learning setting \cite{goodfellow2013empirical, farajtabar2020orthogonal}. In \cite{wen2020batchensemble}, a parameter-efficient ensemble method is described where a rank-one weight matrix is learned per task and used to scale a weight matrix learned on the first task which is shared among ensemble members.

\textbf{Replay}. Replay-based methods are a major category in the taxonomy of approaches to continual learning, and certain replay-based methods are naturally applicable to the task-free setting. For example, in \cite{isele2018selective, rolnick2018experience}, reservoir sampling is used to maintain a uniform sample of all previously seen data in the replay buffer and does not require any knowledge of task boundaries. Other methods mitigate forgetting by curating the contents of the buffer \cite{aljundi2019gradient} or by being selective about which examples to replay \cite{aljundi2019online} in a task-free manner. Certain ``pseudo-rehearsal'' methods that train generative models to mimic past data are also suited to the task-free setting \cite{shin2017continual, kamra2017deep}. Memory-based parameter adaptation \cite{sprechmann2018memorybased} stores a record of past examples like a replay method, but at test-time uses $k$-nearest neighbour look-up to retrieve similar examples to the current input, and uses them to locally adapt network weights. The authors show that a few locally targeted updates restore the nework’s ability to deal with the retrieved exemplars, and hence with the current input. Our architecture does not employ a replay buffer. However, replay methods, such as those cited, are typically compatible with our approach, since our architecture is indifferent to where incoming data comes from. An outer loop that injected examples drawn from a replay buffer, either before or after the pre-trained encoder, would be a straightforward addition.

\textbf{Meta-learning}. Some meta-learning approaches to continual learning do not require the incoming data to be split up into discrete tasks. In \cite{javed2019meta}, a representation-learning network is meta-trained to minimise catastrophic forgetting on sequences of tasks (e.g. split Omniglot) when combined with a prediction network trained in the inner loop. In \cite{beaulieu2020learning}, a neuromodulatory network is meta-trained instead, and used to gate the prediction network. In \cite{he2019task}, a network is meta-trained to infer task representations from incoming data which are then used to condition another network that performs the task at hand. Like our method, these meta-learning approaches involve a kind of pre-training, in the form of the meta-training phase. Other methods exist that are agnostic to the \textit{location} of task boundaries, but nevertheless rely on the fact that the data is piecewise stationary and try to detect task boundaries \cite{kirkpatrick2017overcoming, aljundi2019task}.

\textbf{Gradual distribution change}. A number of other authors have presented continual learning methods that can handle gradual distribution change or soft task boundaries, and have proposed benchmarks akin to our Gaussian schedule to test this \cite{rao2019continual, zeno2018task, lee2020neural}. The CURL approach (Continual Unsupervised Representation Learning) \cite{rao2019continual} is a generative replay method similar to GEN-MIR \cite{aljundi2019online} that can carry out unsupervised clustering in a continual learning setting without knowledge of task boundaries, and can be tailored to supervised learning in a task-free incremental class learning setting as used to evaluate our model. The authors also show that CURL is effective for unsupervised clustering of MNIST digits in the context of a gradually shifting distribution similar to our Gaussian schedule, although they don’t report accuracies for supervised learning in this setting, nor do they apply their model in an online (single epoch) setting, as we do.

\section{Discussion}

We have presented an architecture that combines three straightforward ideas -- a pre-trained encoder, an ensemble of simple classifiers, and a particular activation / loss function pairing -- and shown the effectiveness of this combination in the particularly demanding continual learning setting where task boundaries are either unknown or absent. We have shown that, for a simple a dataset like MNIST, each of these features independently helps to mitigate catastrophic forgetting in this setting, but they are most potent when working together. Used together, this combination of features proves dramatically more effective on harder datasets, such as CIFAR-10 and CIFAR-100, than other methods.

Ensemble methods belong to a family of architectures with a long pedigree in artificial intelligence that begins with Selfridge’s pandemonium architecture \cite{selfridge1959pandemonium} and includes Minsky’s ``society of mind’’ \cite{minsky1988society}, the blackboard systems of the 1980s \cite{nii1986blackboard}, mixture of experts models \cite{jacobs1991adaptive}, and global workspace architecture \cite{shanahan2006cognitive}. All of these architectures feature sets of parallel, independent modules or processes that compete and / or co-operate with each other to collectively determine a system's behaviour. The profound benefits of such architectures -- which are manifest in the biological brain as well as in engineered systems -- can be summarised in three maxims. 1) Competition: selection from a pool of processes or modules encourages specialisation for different contexts, and the right specialist in the right context will outperform a generalist (``jack of all trades, master of none’’). 2) Co-operation: when independent processes or modules specialise in different aspects of a situation, their expertise can be combined in a compositional fashion, which aids generalisation (``divide and rule’’). 3) Collectivity: the aggregated contributions of a diverse set of separate processes or modules yields better performance than any single monolithic process can (``the wisdom of the crowd’’). In this respect, the choice of an ensemble method to deal with catastrophic forgetting is not {\it ad hoc}, but is part of a larger picture wherein modular architectures are used to address some of the deepest problems in building AI that approaches human-level intelligence.

Of course, our method is not without limitations. Chief among these is its reliance on a pre-trained encoder. But recent advances in self-supervised learning have helped to make the case for general-purpose, off-the-shelf encoders that can be applied to datasets different from the one they were trained on. Our work further strengthens this case. For new types of data (for audio, say, or language), a new encoder is needed to make use of our method. Yet if we consider the human capacity for lifelong learning -- something today’s AI systems can only aspire to -- this idea seems natural. If a child is shown a picture book and learns the names of a series of animals it has never seen before, it starts with a perceptual system that has been thoroughly pre-trained on several years of previous visual experience. Similar remarks apply to sound, to touch, to the whole panoply of human experience. Indeed, the ability to build on what has been learned before is the essence of continual learning.

\section*{Code}

Code to accompany this paper is available from \url{https://github.com/deepmind/deepmind-research/tree/master/continual_learning}. This is a Colab notebook, using the JAX library, that works for split MNIST using a pre-trained autoencoder, as described above. The notebook is best run with a GPU to obtain reasonable speed. Running the notebook should reproduce the bottom, right-hand plot of Fig.~\ref{fig:mnist-results}. To get good results for other datasets (eg: CIFAR-10), use an off-the-shelf pre-trained encoder such as BYOL or ReLIC. Code for BYOL is available from \url{https://github.com/deepmind/deepmind-research/tree/master/byol}.

\section*{Acknowledgments}

Thanks to Ben Beyret for help with the open-sourcing process.

\bibliographystyle{plain} 
\bibliography{references}

\cleardoublepage

\appendix

\cleardoublepage

\renewcommand{\thesection}{S\arabic{section}}
\renewcommand{\thetable}{S\arabic{table}}
\renewcommand{\thefigure}{S\arabic{figure}}
\renewcommand{\theequation}{S\arabic{equation}}
\setcounter{equation}{0}

\section{Supplementary Material}

\subsection{Encoder Architecture and Pre-training}

\textbf{MNIST}. For the MNIST experiments, a variational autoencoder (VAE) was pre-trained on the Omniglot training set. The {\em encoder} half of the autoencoder comprised two convolutional layers (kernel size = 4, no. of channels = 16, activation function = ReLU) followed by two heads, one for the encoding means, and the other for the encoding standard deviations. Each head comprised two linear layers. The first linear layer had an output size of 128 with a ReLU activation function, and the second produced the latent encoding with size 512. The final activation function for the means was tanh, and for the standard deviations was ReLU. The {\em decoder} half of the autoencoder was the mirror image of the encoder, comprising two linear layers (with 128 and $28 \times 28 \times 16 = 12844$ output channels, respectively) followed by two transpose convolutional layers (kernel size = 4, no of channels = 16 and 1, respectively, activation function = ReLU and sigmoid, respectively). The autoencoder was trained on 10,000 batches of Omniglot images (batch size = 48), using Adam (learning rate = 0.001), minimising the usual VAE loss function
\begin{align}
    {||x - \hat{x}}||^2 + \beta \mathrm{KL}(\mathcal{N}(\mu_x, \sigma_x), \mathcal{N}(0,1))
\end{align}
where $x$ is the input image, $\hat{x}$ is the output of the decoder given input $z \sim \mathcal{N}(\mu_x, \sigma_x)$, and $(\mu_x, \sigma_x)$ is the output of the encoder given image $x$. We set $\beta = 0.001$. We noted, by visual inspection alone, that the resulting model produced satisfactory reconstructions of both Omniglot and MNIST characters, and didn't attempt to tune the model further. Once trained, the decoder half was discarded, along with the standard deviation head of the encoder. The weights of the remaining portion of the encoder were frozen for subsequent use in the main ensemble model. Prior to each training run of the main model, we pre-trained the autoencoder from scratch (on Omniglot, of course, not MNIST), to ensure statistical variation. Autoencoders were re-trained if the reconstruction loss ${||x - \hat{x}}||^2$ exceeded a certain threshold (0.025).

\textbf{CIFAR-10 and CIFAR-100}. For the CIFAR-10 and CIFAR-100 experiments, we used a ResNet-50 encoder \cite{he2016deep} pre-trained on the ImageNet dataset \cite{deng2009imagenet}. After the pre-training, the weights of the encoder are frozen. The results reported in the main paper were obtained with the ReLIC training method \cite{mitrovic2021representation}. We also obtained results using the BYOL method \cite{grill2020bootstrap} (\ref{table:byol_results}). In both cases, we used the settings and hyperparameters described in the respective original publications cited. In both cases, we upsampled the CIFAR-10 and CIFAR-100 images to a size of $128 \times 128$ using cubic interpolation. The size of the encoder output is 2048.

\begin{figure*}[b]
  \centering
  \includegraphics[width=1.0\textwidth]{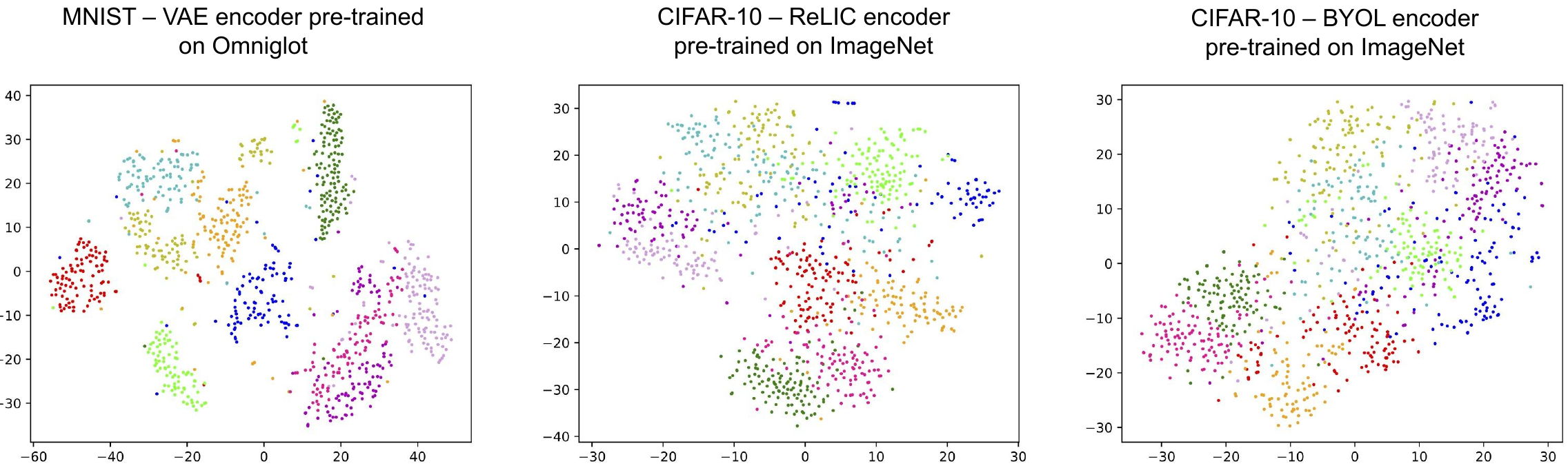}
  \caption{tSNE Plots for image encodings}
  \label{fig:tsne_plots}
\end{figure*}

The success of the model depends on the ability of the pre-trained encoder to map images into a latent space where unseen classes are to a degree linearly separable. Fig.~\ref{fig:tsne_plots} shows tSNE plots for batches of 1000 images from the MNIST and CIFAR-10 datasets after being passed through the relevant pre-trained encoders. In each case, we can see that the encodings exhibit a degree of class-relevant clustering.

\subsection{Ensemble Model, Baselines, and Training}

\textbf{Ensemble model}. Encoded images are passed on to the ensemble for classification. Each classifier in the ensemble comprises a single linear layer followed by a scaled $\mathrm{tanh}$ activation function, as desribed in the main paper. The weights and biases of the classifiers in the ensemble are the model's only trainable parameters. Weights were initialised using fan-in variance scaling with a scaling factor of 1.0 drawn from a truncated normal distribution. Biases were initialised to 0. Each classifier is paired with a unique key, which is used for $k$-nearest neighbour lookup, using cosine similarity, as described in the main paper. The keys, which are fixed throughout training, were drawn from a standard normal distribution. The ensemble model was trained using a ``naive'' optimiser with weight decay, with learning rate 0.0001 and decay factor 0.0001. The naive optimiser works by discarding the magnitude of a gradient having computed its sign, then updating the associated parameter upwards or downwards (according to the sign) by the learning rate. With the exception of ablations, these and all other hyperparameters, were used for all experiments described in the paper (Table~\ref{table:hyperparameters}). We found performance to be robust to hyperparameter variation, and we did not have to tune them for different datasets or experimental settings.

\begin{table}
  \caption{Hyperparameters}
  \label{table:hyperparameters}
  \centering
  \begin{tabular}{ll}
    \toprule
    Parameter                              & Value \\
    \toprule
    Learning rate                          & $0.0001$ \\
    Weight decay                           & $0.0001$ \\
    Ensemble size                          & $1024$ \\
    $k$ (top-$k$ selection)                & $32$ \\
    $\tau$ (tanh scaling factor)           & $250$ \\
    Runs per experiment                    & $20$ \\
    \bottomrule
  \end{tabular}
\end{table}

\textbf{Baselines}. Results for two baseline models are reported throughout the paper: a tanh classifier and a vanilla classifier. These baseline models take the same input as the ensemble mode, namely the output of the pre-trained encoder, but they both consist of a single, stand-alone classifier comprising one linear layer. The only difference between the tanh classifier and the vanilla classifier is their activation functions. The tanh classifier uses a scaled tanh, with the same scaling factor $\tau$ as the classifiers in the ensemble (Table~\ref{table:hyperparameters}, while the vanilla classifier uses a conventional log-softmax function. Weights were initialised using fan-in variance scaling and a truncated normal distribution, as for the ensemble, and biases were initialised to 0. We found that a larger variance scaling factor of 10.0 improved performance for the baselines, so we adopted this. As for the ensemble model, the baselines were trained using a naive optimiser with weight decay, with learning rate 0.0001 and decay factor 0.0001.

\subsection{Experimental Setup and Protocols}

We use standard MNIST, CIFAR-10, and CIFAR-100 datasets \cite{lecun1998gradient, krizhevsky2009learning}. All accuracies were calculated over the full set of class labels (ie: 10 for MNIST and CIFAR-10 and 100 for CIFAR-100). CIFAR-100 accuracies are for the top class label predicted by the model (not top-5, as reported in some papers; our accuracies would be higher if they were top-5). All reported accuracies and all plots in the paper are for the relevant held-out test set. 20 training runs of every experiment were performed, and mean accuracies are reported in tables along with standard deviations, or shown in plots along with error bands of width one standard deviation. For all three datasets, and for every benchmark, we trained the ensemble for exactly 1000 batches. For MNIST, the batch size was five (low data regime) or 60 (otherwise). For CIFAR-10 and CIFAR-100, the batch size was 48. Only one pass through the dataset (one epoch) was performed, corresponding to the ``online’’ setting. (To allow for schedules with gradual distribution shift, every batch was drawn at random from the dataset, after shuffling. So the model is likely to have seen a number of images more than once. However, the total number of images seen in all cases was less than or equal to the size of the training set, in other words a single epoch.)

\textbf{5-way split MNIST}. The ten class labels were partitioned at random into a sequence of five subsets (or tasks). Each training run was carried out on a new random partitioning of labels. (We note that some authors report results for where the partitioning is the same in every run, usually [{0, 1}, {2, 3}, ...].) In each training run, the five tasks were presented sequentially, each task comprising 200 batches randomly drawn from the relevant subset of the training set. The total number of images seen is $5 \times 200 \times 60 = 60000$, which is equivalent to a single pass through the data (one epoch).

\textbf{5-way split MNIST (low data)}. The protocol for this benchmark is identical to that for 5-way split MNIST, but with a smaller batch size (5). The total number of images seen is $5 \times 200 \times 5 = 5000$, which is equivalent to a $1/12$ of a pass through the data ($1/12$ of an epoch).

\textbf{10-way split MNIST}. This is the ``fully incremental’’ benchmark, where the classes are learned one at a time. The ten class labels were partitioned into a sequence of ten singleton subsets (or tasks), each containing just one label. Each training run was carried out on a new random partitioning of labels (ie: a new random ordering of the digits). In each training run, the ten tasks were presented sequentially, each task comprising 100 batches randomly drawn from the relevant subset of the training set. The total number of images seen is $10 \times 100 \times 60 = 60000$, which is equivalent to a single pass through the data (one epoch).

\begin{figure*}
  \centering
  \includegraphics[width=0.9\textwidth]{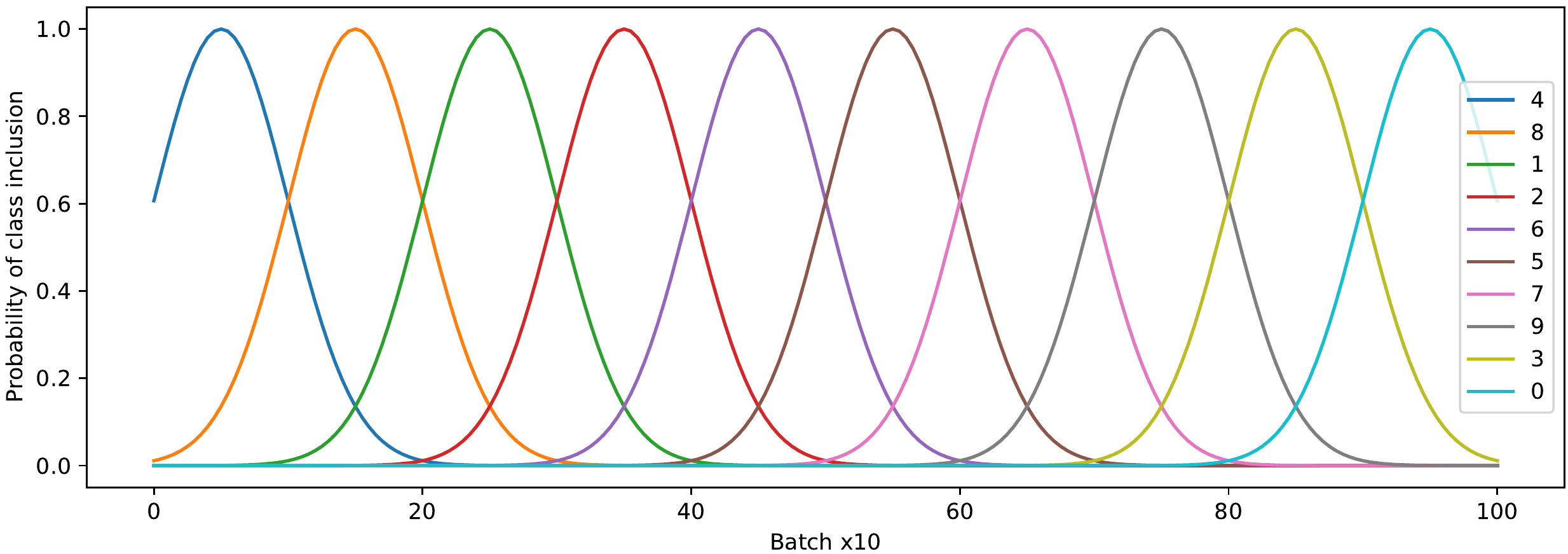}
  \caption{Gaussian schedule}
  \label{fig:gaussian_schedule}
\end{figure*}

\textbf{Gaussian schedule MNIST}. In this benchmark, the data distribution shifts gradually as training proceeds. A series of 200 ``micro-tasks’’ $T_1$ to $T_{200}$ were presented to the model, each comprising 5 batches (of 60 images each). Each micro-task $T_i$ was associated with a random set of class labels $L(T_i)$, and all images in that micro-task were drawn at random from the corresponding subset of the training set. The composition of $L(T_i)$ was determined by the following algorithm, where $m=10$ classes (see Fig.~\ref{fig:gaussian_schedule}). The $m$ class labels are partitioned into a sequence of $m$ singleton subsets $C_1$ to $C_{10}$, each containing just one label. (Fig.~\ref{fig:gaussian_schedule} illustrates one such partitioning.) Each of the ten classes $C_i$ to $C_{10}$ is then associated with a conditional probability distribution $P(C_i | B)$, where $B$ is a batch number. $P(C_i | B)$ is given by a Gaussian function of height 1 centred at $10i$ with width 50. In other words,
\begin{align}
P(C_i | B) = h.e^{(-\frac{(B - 10i)}{2w^2})}
\end{align}
where $h=1$ and $w=50$. The total number of images seen is $100 \times 10 \times 60 = 60000$, which is equivalent to a single pass through the data (one epoch).

\textbf{5-way split CIFAR-10}. The protocol for this benchmark is identical to that of 5-way split MNIST. The total number of images seen is $5 \times 200 \times 48 = 48000$, which is equivalent to a slightly less than a single pass through the data (less than one epoch). (We chose a batch size of 48 rather than 50 for a fair comparison with \cite{aljundi2019online}, who train on 48750 images for split CIFAR-10.)

\textbf{10-way split CIFAR-10}. The protocol for this benchmark is identical to that of 10-way split MNIST. The total number of images seen is $10 \times 100 \times 48 = 48000$, which is equivalent to a slightly less than a single pass through the data (less than one epoch).

\textbf{Gaussian schedule CIFAR-10}. The protocol for this benchmark is identical to that of Gaussian schedule MNIST. The total number of images seen is $100 \times 10 \times 48 = 48000$, which is equivalent to a slightly less than a single pass through the data (less than one epoch).

\textbf{20-way split CIFAR-100}. In the CIFAR-100 dataset each image has an associated fine class label (one of 100) and super-class label (one of 20). If two images belong to the same fine class, then they must belong to the same super-class. For the 20-way split CIFAR-100 benchmark, the 100 fine class labels were partitioned into a randomly ordered sequence of 20 subsets (tasks), with each subset corresponding to one of the super-classes. In each training run, the 20 tasks were presented sequentially, with each task comprising 50 batches (batch size 48) randomly drawn from the relevant subset of the training set.

\textbf{100-way split CIFAR-100}. This is the ``fully incremental’’ benchmark, where the classes are learned one at a time. The 100 fine class labels were partitioned into a sequence of 100 singleton subsets (or tasks), each containing just one label. Each training run was carried out on a new random partitioning of labels (ie: a new random ordering of the digits). In each training run, the 100 tasks were presented sequentially, each task comprising 10 batches (batch size 48) randomly drawn from the relevant subset of the training set.

\textbf{Gaussian schedule CIFAR-100}. The protocol for this benchmark is the same as for Gaussian MNIST and Gaussian CIFAR-10, and uses the same algorithm to generate a schedule of micro-tasks but with $m=100$.

We also evaluated the model (and baselines) for each dataset in the i.i.d. setting (Table~\ref{table:iid_accuracies}). The setup here is the same as for all the other benchmarks: the images are fed through the pre-trained encoder and the resulting encodings are passed on the the ensemble (or baseline classifier). However, each batch is drawn uniformly from the set of all classes.

\subsection{Ablations and Other Experiments}

\begin{table}
  \caption{Ablations. Left: decreasing the ensemble size results in worse performance. Right: increasing $k$ in top-$k$ classifier selection results in worse performance (ensemble size is 1024). All results are final accuracy (\%) over 20 runs for all ten classes.}
  \label{table:ablations}
  
  \vspace{0.2cm}
  
  \centering
    \includegraphics[width=1.0\textwidth]{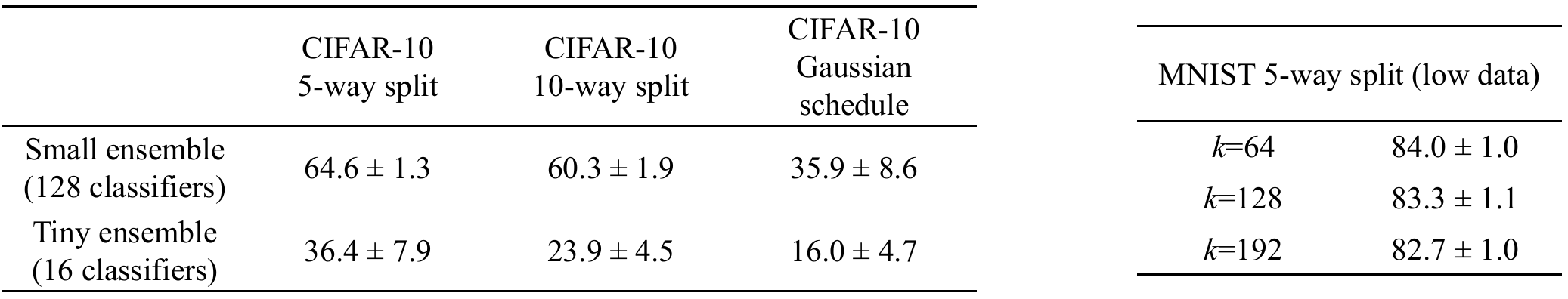}
\end{table}

\begin{table}
  \caption{BYOL encoder results.}
  \label{table:byol_results}
  
  \vspace{0.2cm}
  
  \centering
    \includegraphics[width=0.7\textwidth]{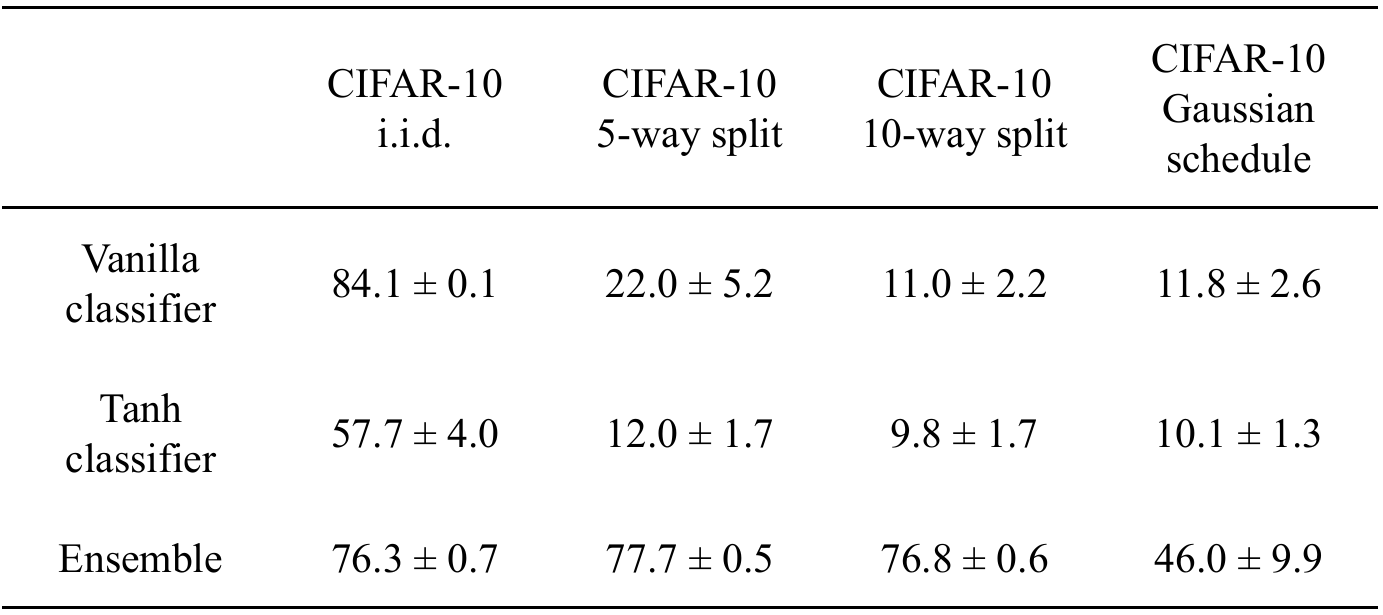}
\end{table}

We carried out a number of ablations, as described in the main text, relating to ensemble size and $k$ in top-$k$ lookup. The results are shown in Table~\ref{table:ablations}. We also carried out the CIFAR-10 and CIFAR-100 experiments using the BYOL encoder rather than ReLIC, yielding the results in Table~\ref{table:byol_results}. Table~\ref{table:iid_accuracies} shows accuracies for all benchmarks using the i.i.d. protocol. Finally, Fig.~\ref{fig:tsne_plots} shows tSNE plots illustrating the clustering of the encoded images with each type of pre-trained encoder we used.

\begin{table}[t!]
  \caption{i.i.d. accuracies.}
  \label{table:iid_accuracies}
  
  \vspace{0.2cm}
  
  \centering
  \includegraphics[width=0.6\textwidth]{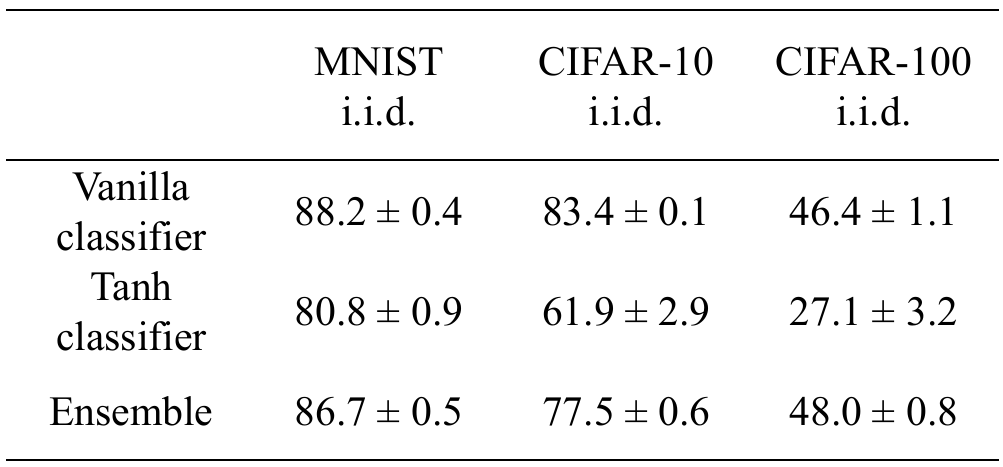}
  \vspace{17cm}
\end{table}

\end{document}